\documentclass[conference]{IEEEtran}
\IEEEoverridecommandlockouts
\usepackage{cite}
\usepackage{amsmath,amssymb,amsfonts}
\usepackage{graphicx}
\usepackage{textcomp}
\usepackage{url}
\usepackage{hyperref}
\usepackage{xcolor}
\usepackage{booktabs}
\usepackage{listings}
\usepackage{microtype}
\usepackage{balance}
\usepackage{caption}
\def\BibTeX{{\rm B\kern-.05em{\sc i\kern-.025em b}\kern-.08em
    T\kern-.1667em\lower.7ex\hbox{E}\kern-.125emX}}

\begin{document}

\title{Cut Binary Cross Entropy: Efficient Large-Vocabulary Loss and Gradient Kernels for Sequential Recommendation}

\author{\IEEEauthorblockN{Yaoyiran Li}
\IEEEauthorblockA{\textit{Google Cloud} \\
London, United Kingdom \\
yaoyiran@google.com}
\and
\IEEEauthorblockN{Haowen Ning}
\IEEEauthorblockA{\textit{Google Cloud} \\
Mountain View, USA \\
hning@google.com}
\and
\IEEEauthorblockN{Mohamed Hammad}
\IEEEauthorblockA{\textit{Google Cloud} \\
Mountain View, USA \\
mibra@google.com}
}

\maketitle

\begin{abstract}
Industrial sequential recommender systems operate over massive item catalogs (e.g., $10^5$--$10^7$ items). Multi-label recommendation models are trained with Binary Cross-Entropy (BCE) loss over the full vocabulary, but standard BCE materializes a dense $[B, N, V]$ logits tensor in High Bandwidth Memory (HBM), incurring prohibitive $O(BNV)$ memory and fatal Out-Of-Memory (OOM) errors. While chunked loss optimizations exist for Softmax Cross-Entropy in LLMs, large-scale multi-label BCE optimization remains unexplored across deep learning ecosystems.

We propose \textbf{CutBCE}, an exact, hardware-accelerated BCE loss and gradient operator implemented in \textbf{JAX} and \textbf{Pallas} for large-vocabulary workloads. CutBCE introduces (1) an exact fused reformulation evaluating dense background loss and sparse target corrections; (2) a custom Vector-Jacobian Product (VJP) with a dedicated Pallas TPU backward kernel computing logit tiles on-chip in both passes so logits and their gradients never reside in HBM; (3) dynamic VMEM budgeting and sharding-aware collective hoisting for distributed meshes; and (4) count-based zero-overhead training metrics. On single-chip TPU v5e/v6e mini-benchmarks, CutBCE eliminates OOM errors with up to 91.9\% speedup. On 8-chip TPU slice training for multi-label SASRec with 876k items (Yambda-50M), CutBCE reduces peak HBM by 65.7\% ($>14$~GiB saved per chip) and increases training speed by 225.9\% with comparable accuracy. CutBCE is open-sourced at \href{https://github.com/AI-Hypercomputer/RecML/blob/main/recml/core/ops/binary_cross_entropy_ops.py}{\nolinkurl{https://github.com/AI-Hypercomputer/RecML/blob/main/recml/core/ops/binary_cross_entropy_ops.py}}.
\end{abstract}


\section{Introduction and Motivation}

Sequential recommender systems are foundational to modern digital services, including e-commerce marketplaces, video streaming platforms, and app stores. These systems model user interaction trajectories to predict future user actions over massive item catalogs containing hundreds of thousands to tens of millions of item IDs~\cite{b1,b3,b4}. Although text-based generative Large Language Models (LLMs) have demonstrated strong semantic reasoning capabilities for recommendation~\cite{b2}, ID-based sequential models (such as SASRec~\cite{b3} and BERT4Rec~\cite{b4}) remain the industry mainstream due to sub-millisecond serving latency via Approximate Nearest Neighbor (ANN) search and lightweight model architectures.

In industrial sequential recommendation, users frequently interact with multiple related items in a single session. Because single-step next-item prediction is myopic and overfits to noisy clicks and arbitrary local order, models are increasingly trained to predict the next $L$ items (e.g., $L=8$) simultaneously~\cite{b11,b12}. This multi-label objective provides richer supervision and captures multi-intent sessions. The standard training loss for multi-label classification is Binary Cross-Entropy (BCE) evaluated over the entire item vocabulary ($V$ is the vocab size).

\subsection{The Logits Memory Wall in Standard BCE}

Given batch size $B$, sequence length $N$, and hidden dimension $D$, the transformer backbone outputs activations $\mathbf{A} \in \mathbb{R}^{B \times N \times D}$. To compute BCE loss, these activations are projected against an item embedding matrix $\mathbf{E} \in \mathbb{R}^{V \times D}$ to form a dense logits tensor: $\mathbf{X} = \mathbf{A} \mathbf{E}^T \in \mathbb{R}^{B \times N \times V}$. When vocabulary size $V$ reaches industrial scales (e.g., $V = 876,939$ in the Yambda dataset~\cite{b1}, comprising 876,938 items and one padding token), materializing $\mathbf{X}$ creates a severe memory wall.

In distributed data-parallel training across an 8-chip TPU slice ($2\times 4$ topology), even with a moderate local batch size $B_{\text{local}} = 16$ (global batch $B=128$), sequence length $N=256$, and vocab size $V=876,939$, storing the single local logits tensor $\mathbf{X}_{\text{local}} \in \mathbb{R}^{16 \times 256 \times 876,939}$ in 32-bit float requires $16 \times 256 \times 876,939 \times 4\text{ B} \approx 13.38\text{ GiB}$. While this alone would consume nearly $83.6\%$ of a 16~GiB accelerator's memory (e.g., TPU v5e), even on a 32~GiB TPU v6e (Trillium), allocating 13.38~GiB solely for logits leaves insufficient HBM for model parameters, embeddings, optimizer states, transformer activations, and compiler buffers, triggering fatal Out-Of-Memory (OOM) crashes at $B=128, N=256$ (as observed in Table~\ref{tab:sasrec}). Furthermore, standard autodiff in JAX~\cite{b8}, PyTorch, and Keras~\cite{b7} retains $\mathbf{X}$ in HBM across forward and backward passes (alongside dense targets and $\partial\mathcal{L}/\partial\mathbf{X}$), overlapping directly with transformer activations.

\subsection{Research Gap and Key Contributions}

To mitigate logit materialization bottlenecks, recent works have explored chunked loss operators, such as Cut Cross Entropy (CCE)~\cite{b5}, Liger Kernel~\cite{b10}, and Tokamax~\cite{b6}. However, existing efforts suffer from two key limitations:
\begin{enumerate}
\item \textbf{Exclusive Focus on Softmax Cross-Entropy}: Prior optimizations target Softmax Categorical Cross-Entropy (CE) for auto-regressive LLMs with global partition functions ($\log \sum_{v=1}^V e^{x_v}$). In contrast, BCE evaluates independent Bernoulli negative log-likelihoods per class with sparse positive targets (where the number of positive labels $L \ll V$). BCE efficiency optimization for large vocabularies remains unexplored across deep learning ecosystems (e.g., PyTorch/GPU and JAX/TPU).
\item \textbf{PyTorch and GPU Centricity}: Prior loss kernels (e.g., Liger Kernel) rely on CUDA/Triton for Nvidia GPUs. In Google Cloud infrastructure, large-scale recommendation training is predominantly powered by JAX~\cite{b8} on Tensor Processing Units (TPUs). Developing efficient, distributed, sharding-aware custom loss kernels in JAX and Pallas~\cite{b9} presents distinct architectural requirements.
\end{enumerate}

To bridge this gap, this paper introduces \textbf{CutBCE}, an exact, hardware-accelerated BCE loss operator implemented in JAX and Pallas. The key contributions of this work are as follows:
\begin{itemize}
\item \textbf{Fused Direct BCE Formulation}: Exact mathematical decomposition of BCE loss into dense background and sparse target correction.
\item \textbf{Custom VJP \& Pallas TPU Kernel}: A hybrid JAX/XLA forward scan with a dedicated Pallas TPU backward kernel acting as an explicit rematerialization mechanism that never materializes full logits in HBM.
\item \textbf{Hardware-Aware Distributed Optimizations}: Dynamic VMEM budgeting, hierarchical sequence chunking, and sharding-aware collective hoisting across device meshes.
\item \textbf{Zero-Overhead Cut Metrics}: Linear count-based confusion matrix computation directly during the forward pass with zero auxiliary HBM overhead.
\item \textbf{Empirical Demonstration}: Rigorous validation on synthetic benchmarks and multi-label SASRec on the 876k-item Yambda dataset~\cite{b1}, demonstrating 65.7\% HBM reduction and 225.9\% speedup with comparable accuracy.
\end{itemize}

\section{Related Work}

\noindent\textbf{Loss Optimization for Large Vocabularies:} As vocabulary sizes scale, output projection and loss computation dominate training memory and compute budgets. While approximate methods like negative sampling reduce computation by evaluating only a subset of negatives, they introduce sampling bias and alter the optimization objective. For exact full-vocabulary training, Cut Cross Entropy (CCE)~\cite{b5} pioneered vocabulary-dimension chunking for LLM cross-entropy. In the GPU ecosystem, Liger Kernel~\cite{b10} implemented Triton-based fused kernels in PyTorch for LLM fine-tuning. In the TPU ecosystem, Tokamax~\cite{b6} provides Pallas-based linear softmax cross-entropy for PyTorch. However, all existing chunked loss frameworks exclusively target \textit{Softmax Categorical Cross-Entropy} with global normalization. In contrast, multi-label recommendation evaluates independent Bernoulli negative log-likelihoods per class over the whole item vocabulary ($L \ll V$). To the best of our knowledge, CutBCE is the first work (in both JAX/TPU and PyTorch/GPU) to optimize and hardware-accelerate exact BCE loss and gradients for large vocabularies.


\noindent\textbf{JAX, Pallas, and Accelerator Kernels:} 
JAX~\cite{b8} compiles numerical programs via XLA and provides functional transformations such as \texttt{custom\_vjp} and \texttt{shard\_map}. For fine-grained control over accelerator memory hierarchies, Pallas~\cite{b9}, JAX's kernel language, offers direct control over Vector Memory (VMEM), Matrix Multiply Units (MXUs), Vector Processing Units (VPUs), and DMA transfers on Google Cloud TPUs. While Pallas has been adopted for, e.g., attention and softmax CE kernels, Pallas kernels for large-vocabulary BCE have been absent. CutBCE fills this gap with a custom Pallas backward kernel optimized for TPU scratchpad hierarchies.

\noindent\textbf{Sequential Recommendation Systems:} Sequential recommender systems model user interaction trajectories over time. Self-attentive architectures (SASRec~\cite{b3} and BERT4Rec~\cite{b4}) and LLM alignment methods (CALRec~\cite{b2}) predict future user engagement over catalogs with $10^5$--$10^7$ item IDs. In industrial applications, multi-intent sessions motivate training with multi-label objectives~\cite{b11,b12}, making fast, memory-efficient BCE computation in JAX/Pallas crucial for scalable training.

\section{Methodology}

\subsection{Mathematical Formulation of Fused Direct BCE}

Let $\mathbf{a} \in \mathbb{R}^D$ denote a model's hidden representation at a sequence position, and $\mathbf{E} \in \mathbb{R}^{V \times D}$ denote the item output embedding matrix. The logit for class $v \in \{1, \dots, V\}$ is $x_v = \mathbf{a} \cdot \mathbf{e}_v^T$, and $y_v \in \{0, 1\}$ denotes the binary target. The standard BCE loss for predicted probability $p_v = \sigma(x_v) = \frac{1}{1 + e^{-x_v}}$ is
\begin{equation}
\mathcal{L}(x_v, y_v) = - y_v \log(\sigma(x_v)) - (1 - y_v) \log(1 - \sigma(x_v)).
\label{eq:bce_expanded}
\end{equation}
Substituting $\sigma(x_v)$ into \eqref{eq:bce_expanded} yields $\mathcal{L}(x_v, y_v) = x_v + \log(1 + e^{-x_v}) - y_v x_v$, but the term $\log(1 + e^{-x})$ is numerically unstable for large negative $x$ as $e^{-x}$ overflows floating-point range.
Using the log-sum-exp identity $x + \log(1 + e^{-x}) = \max(x, 0) + \log(1 + e^{-|x|})$, we obtain the numerically stable formulation:
\begin{align}
\mathcal{L}(x_v, y_v) &= \max(x_v, 0) - y_v x_v + \log(1 + e^{-|x_v|}) \nonumber \\
&= \mathcal{L}(x_v, 0) - y_v \cdot x_v,
\label{eq:fused_decomposition}
\end{align}
where $\mathcal{L}(x_v, 0) = \max(x_v, 0) + \log(1 + e^{-|x_v|})$ is the dense background loss, and $y_v \cdot x_v$ is the sparse target correction (active only for positive targets $y_v = 1$). The total loss over the whole item vocabulary is defined as follows:
\begin{equation}
\mathcal{L}_{\text{total}} = \frac{1}{V} \sum_{v=1}^V \Big[ \max(x_v, 0) + \log(1 + e^{-|x_v|}) - y_v \cdot x_v \Big].
\label{eq:total_loss}
\end{equation}

\noindent\textbf{Vocabulary Chunking Algorithm:} Rather than computing $\mathbf{X} \in \mathbb{R}^{B \times N \times V}$ globally, CutBCE tiles the vocabulary into blocks of size $B_V$ (\texttt{block\_v}, e.g., $B_V = 4096$).\footnote{In practice, CutBCE flattens the batch and sequence dimensions into a single $BN$ dimension and hierarchically tiles it (via sequence chunking and inner kernel tiles of size $B_N=128$) to respect accelerator VMEM capacity.} For each block $j \in \{0, \dots, \lceil V/B_V \rceil - 1\}$: (1) slice $\mathbf{E}_j \in \mathbb{R}^{B_V \times D}$; (2) compute chunked logits $\mathbf{X}_j = \mathbf{A} \mathbf{E}_j^T \in \mathbb{R}^{B \times N \times B_V}$; (3) accumulate the block's loss into a running sum; and (4) discard $\mathbf{X}_j$ immediately. This reduces peak intermediate activation memory from $O(BNV)$ to $O(BNB_V)$, achieving significant memory reductions.

\subsection{Gradient Formulation and Custom VJP Rematerialization}

The analytical gradient of the BCE loss with respect to logit $x_v$ is $\frac{\partial \mathcal{L}}{\partial x_v} = \frac{1}{V} (\sigma(x_v) - y_v)$. Let $\mathbf{Y}$ denote the multi-hot target tensor. The gradients for activations $\mathbf{A}$ and embeddings $\mathbf{E}$ are given as follows:
\begin{equation}
\nabla_{\mathbf{A}} \mathcal{L} = \frac{1}{V} (\sigma(\mathbf{X}) - \mathbf{Y}) \mathbf{E}, \quad \nabla_{\mathbf{E}} \mathcal{L} = \frac{1}{V} (\sigma(\mathbf{X}) - \mathbf{Y})^T \mathbf{A},
\label{eq:matrix_grads}
\end{equation}
where $\mathbf{A}$, $\mathbf{X}$, and $\mathbf{Y}$ are viewed as 2D matrices with $B$ and $N$ flattened into $BN$ rows. Averaging the loss over the $BN$ positions adds a factor of $1/(BN)$ to both; in practice, padding positions are excluded from this average.

\begin{table}[t]
\caption{Logit Memory Comparison.}
\label{tab:lifespan}
\centering
\scriptsize
\setlength{\tabcolsep}{3pt}
\begin{tabular}{lll}
\toprule
\textbf{Memory Footprint} & \textbf{Standard BCE} & \textbf{CutBCE} \\
\midrule
Logits & $[B, N, V]$ & $[BN, B_V]$ on-chip tile \\
Logit Gradient & $[B, N, V]$ & $[B_N, B_V]$ on-chip tile \\
Complexity & $O(B N V)$ & $O(B N B_V)$ \\
\bottomrule
\end{tabular}
\end{table}

\noindent\textbf{Breaking Memory Overlap via Custom VJP:} In standard autodiff, the forward pass records intermediate activations onto a reverse tape, keeping the full $[B, N, V]$ logits resident in HBM across forward and backward passes. CutBCE defines a custom Vector-Jacobian Product (\texttt{custom\_vjp}). In the forward pass, only $\mathbf{A}$, $\mathbf{E}$, and sparse $[B, N, L]$ target IDs ($L \ll V$) are retained; no multi-hot target tensor is materialized. During backward pass, the custom kernel recomputes logits $\mathbf{X}_j$ chunk-by-chunk on the fly and accumulates gradients directly. As shown in Table~\ref{tab:lifespan}, no $[B, N, V]$ buffer is materialized in HBM, drastically lowering global peak memory.

\subsection{Hybrid Architecture and Pallas TPU Backward Kernel}

We engineer a \textbf{Hybrid JAX/XLA Forward + Pallas TPU Backward} architecture. The backward pass executes via a dedicated Pallas TPU kernel (\texttt{\_bce\_bwd\_kernel}):
\begin{itemize}
\item \textbf{Local VMEM Scratchpad Accumulation}: Embedding gradients $\nabla_{\mathbf{E}_j}\mathcal{L}$ are accumulated in TPU VMEM across inner kernel tiles ($B_N = 128$, matching the TPU vector lane width) before a single contiguous DMA write back to HBM, drastically reducing memory bus traffic.
\item \textbf{Dynamic VMEM Budgeting \& Sequence Chunking}: CutBCE dynamically queries physical VMEM capacity via TPU device info. When the total sequence dimension $BN$ exceeds safe scratchpad limits, the chunking dispatcher (\texttt{\_bce\_bwd\_pallas\_\allowbreak chunked\_n}) partitions $BN$ into outer sequence chunks processed sequentially in HBM.

\item \textbf{Sharding-Aware Collective Hoisting}: For distributed meshes (Data, FSDP, or Vocab Parallelism), dispatcher \texttt{\_bce\_bwd\_sharded} leverages \texttt{jax.shard\_map} to hoist weight replication constraints outside the loop, manage per-shard \texttt{vocab\_offset}, and perform collective \texttt{psum} reductions across mesh axes.
\end{itemize}

\begin{lstlisting}[language=Python,caption={CutBCE drop-in usage in JAX/Keras 3.}]
from recml.core.ops import binary_cross_entropy_ops

# activations: [B,N,D], embeddings: [V,D], targets: [B,N,L]
loss = binary_cross_entropy_ops.cut_binary_cross_entropy(
    activations=activations, embeddings=embeddings,
    targets=targets, weights=weights, block_v=4096,
    mesh=mesh, act_spec=act_spec, emb_spec=emb_spec)
\end{lstlisting}

\subsection{Zero-Overhead Cut Metrics for Training}

Standard count-based metrics reconstruct the full $[B, N, V]$ logits during training, defeating memory savings. CutBCE introduces \textbf{Cut Metrics}: during the forward pass, it computes TP, FP, FN, and TN counts of shape $[B, N]$ without materializing $[B, N, V]$ logits. Standard classification metrics (such as binary accuracy, precision, recall, and $F_\beta$) can then be calculated directly from these aggregated counts with zero auxiliary HBM overhead.

\noindent\textbf{Training vs. Evaluation Metrics:} Cut Metrics are designed exclusively for monitoring classification metrics during the \textit{training phase} without logit reconstruction. In contrast, ranking-based evaluation metrics (e.g., our multi-label $\text{Hit}@8$, Section~\ref{sec:sasrec_pipeline}) are evaluated during separate evaluation/test phases over full logits, completely decoupled from Cut Metrics.

\subsection{Multi-Label SASRec-BCE Application Pipeline}
\label{sec:sasrec_pipeline}

We implement a causal, non-autoregressive multi-label sequential recommendation pipeline based on SASRec~\cite{b3} with a GPT-2 transformer backbone.

\noindent\textbf{Task Formulation:} The model predicts the next $L=8$ items simultaneously. Input and target item IDs lie in $\mathbb{Z}^{B \times N}$ and $\mathbb{Z}^{B \times N \times L}$, respectively. For a user sequence of length $n$: (1) the \textit{test set} uses the first $n-8$ items as input and the last 8 items as targets; (2) the \textit{dev set} uses the first $n-17$ items as input and the following 8 items as targets; and (3) the \textit{training set} uses the first $n-26$ items as input, where every non-padding position is trained via CutBCE to predict its next 8 items. In all splits, the input is truncated to at most the $N$ most recent items and padded to length $N$. At evaluation, the hidden state at each user's last non-padding position is projected to full-vocabulary logits ($\mathbb{R}^V$, no negative sampling), and we report multi-label Hit@8: a user counts as a hit if at least one of the 8 held-out next items appears among the top-8 ranked items.

\section{Experimental Evaluation}
\label{experiments}
\noindent\textbf{Hardware Setup:} Experiments are run on Google Cloud \href{https://docs.cloud.google.com/tpu/docs/v6e}{TPU v6e} (Trillium, 32~GiB HBM) and \href{https://docs.cloud.google.com/tpu/docs/v5e}{TPU v5e} (16~GiB HBM) with JAX and Keras 3.

\subsection{Mini-Benchmark Experiments}
\label{experiments_a}
\balance
We benchmark isolated loss + gradient step time on a \textbf{single TPU chip} (TPU v6e/v5e) across 500 steady-state iterations after JIT compilation. Here, ``isolated loss + gradient'' measures the loss and backward gradient computation in isolation without transformer backbone or optimizer overhead. Input tensors are generated synthetically: $\mathbf{A} \sim \mathcal{N}(0, \mathbf{I})$ of shape $[B, N, D]$, $\mathbf{E} \sim \mathcal{N}(0, \mathbf{I})$ of shape $[V, D]$, and target labels sampled uniformly from $[0, V)$ of shape $[B, N, L]$, with hidden dimension $D=128$ and $L=8$ positive labels per position.

\noindent\textbf{Numerical Equivalence:} At $B=N=32$ and $V=100$k, CutBCE matches standard Keras BCE on both TPU v5e and v6e within $1.1\times10^{-7}$ relative loss error and $7.1\times10^{-5}$ relative $\ell_2$ error in $\nabla_{\mathbf{A}}\mathcal{L}$ and $\nabla_{\mathbf{E}}\mathcal{L}$, i.e., negligible numerical differences. Our open-source unit tests further check CutBCE against Keras across unaligned $N$ and $V$, padded and duplicate target IDs, and Pallas and pure-JAX backends.

\begin{table}[t]
\caption{Mini-Benchmark: Standard Keras BCE vs. CutBCE on a Single TPU v6e/v5e ($D=128, L=8$). Step times averaged over 500 iterations.}
\label{tab:minibench}
\centering
\scriptsize
\setlength{\tabcolsep}{3pt}
\renewcommand{\arraystretch}{0.91}
\begin{tabular}{lllll}
\toprule
\textbf{Batch $\times$ SeqLen} & \textbf{Vocab ($V$)} & \textbf{Keras (ms)} & \textbf{CutBCE (ms)} & \textbf{Speedup} \\
\midrule
\multicolumn{5}{c}{\textit{Google Cloud TPU v6e (Trillium, 32 GiB HBM, Single Chip)}} \\
$128 \times 128$ & 100,000 & OOM & \textbf{6.93} & -- \\
$32 \times 32$   & 100,000 & 1.43 & \textbf{0.93} & +53.8\% \\
$32 \times 32$   & 200,000 & 2.86 & \textbf{1.49} & +91.9\% \\
\midrule
\multicolumn{5}{c}{\textit{Google Cloud TPU v5e (16 GiB HBM, Single Chip)}} \\
$128 \times 128$ & 100,000 & OOM & \textbf{12.79} & -- \\
$32 \times 32$   & 100,000 & 2.49 & \textbf{1.47} & +69.4\% \\
$32 \times 32$   & 200,000 & 4.62 & \textbf{2.52} & +83.3\% \\
\bottomrule
\end{tabular}
\end{table}

\begin{table}[t]
\caption{End-to-End Multi-Label SASRec on Yambda-50M ($V=876,939$) on an 8-chip TPU v6e slice ($2\times 4$ Topology, 32\,GiB HBM/chip).}
\label{tab:sasrec}
\centering
\scriptsize
\setlength{\tabcolsep}{2.5pt}
\renewcommand{\arraystretch}{0.91}
\begin{tabular}{lllll}
\toprule
\textbf{Batch $\times$ SeqLen} & \textbf{Method} & \textbf{HBM (GiB)} & \textbf{Speed (steps/s)} & \textbf{Hit@8} \\
\midrule
$64 \times 256$  & SASRec-BCE    & 22.37 & 6.14 & 0.186 \\
$64 \times 256$  & SASRec-CutBCE & \textbf{7.67} (\textit{-65.7\%}) & \textbf{20.01} (\textit{+225.9\%}) & 0.183 \\
\midrule
$128 \times 256$ & SASRec-BCE    & OOM   & --   & -- \\
$128 \times 256$ & SASRec-CutBCE & \textbf{7.89} & \textbf{15.82} & 0.166 \\
\midrule
$64 \times 512$ & SASRec-BCE    & OOM   & --   & -- \\
$64 \times 512$ & SASRec-CutBCE & \textbf{7.89} & \textbf{15.61} & 0.192 \\
\midrule
$128 \times 512$ & SASRec-BCE    & OOM   & --   & -- \\
$128 \times 512$ & SASRec-CutBCE & \textbf{8.94} & \textbf{10.74} & 0.211 \\
\bottomrule
\end{tabular}
\end{table}

As shown in Table~\ref{tab:minibench}: (1) \textbf{OOM Elimination}: At $B=128, N=128, V=100\text{k}$, standard Keras BCE crashes with OOM, while CutBCE executes smoothly in 6.93~ms (TPU v6e) and 12.79~ms (TPU v5e). (2) \textbf{Speedups}: At $B=32, N=32$, CutBCE delivers 53.8\%--91.9\% speedups on TPU v6e and 69.4\%--83.3\% on TPU v5e. (3) \textbf{Vocabulary Scaling}: Speedup improves as vocabulary grows due to high VMEM cache locality and reduced HBM bandwidth roundtrips.

\subsection{Multi-Label SASRec Experiments on Yambda Dataset}
\label{experiments_b}
We evaluate end-to-end training on \textbf{Yambda-50M}~\cite{b1}, filtering out users with $\le 30$ interactions (retaining 9,074 users, 46.46M interactions, average sequence length 5,120.56, 876,938 unique items). The model uses a 3-layer GPT-2 backbone (4 attention heads, hidden dim $D=256$) and is trained for $300,000$ steps with AdamW (lr=$5e-4$, $5,000$ warm-up steps). 

\noindent\textbf{Distributed Setup:} Training is conducted on an \textbf{8-chip TPU v6e slice} ($2\times 4$ topology) using Data Parallelism (\texttt{keras.distribution.DataParallel}), where each chip receives $1/8$ of the global batch (local batch $B_{\text{local}} = 8$ for global $B=64$, and $B_{\text{local}} = 16$ for global $B=128$). Validation (dev) and test evaluation pipelines run concurrently on dedicated, physically independent TPU slices. Training, dev, and test execution never share or contend for TPU hardware.

As shown in Table~\ref{tab:sasrec}: (1) \textbf{65.7\% Training HBM Reduction}: At global $B=64, N=256$, SASRec-BCE consumes 22.37~GiB HBM per chip, whereas SASRec-CutBCE requires only 7.67~GiB HBM (saving 14.70~GiB). (2) \textbf{225.9\% Speedup}: Training speed increases from 6.14 to 20.01 steps/second. (3) \textbf{Larger Batch and Seq Length}: At global $B=128$ or $N=512$, standard BCE crashes with OOM, while CutBCE runs stably at $7.89$--$8.94$~GiB HBM. (4) \textbf{Meet Accuracy Target}: CutBCE achieves 0.183 Hit@8 vs 0.186 for standard BCE.

\section{Conclusion and Open Source}
We presented CutBCE, an exact, memory-efficient BCE operator for large-vocabulary sequential recommendation featuring fused direct BCE formulation, custom VJP rematerialization with Pallas TPU backward kernels, sharding-aware collective hoisting, and count-based metrics. It achieves up to 225.9\% training speedups, 65.7\% HBM reduction, and unblocks large-batch, long-sequence training on 876k-item catalogs. CutBCE is open-sourced as a modular JAX operator in RecML at \href{https://github.com/AI-Hypercomputer/RecML/blob/main/recml/core/ops/binary_cross_entropy_ops.py}{\nolinkurl{https://github.com/AI-Hypercomputer/RecML/blob/main/recml/core/ops/binary_cross_entropy_ops.py}}. It serves as a drop-in replacement for Keras and JAX models, handling tile sizes, sharding, and Pallas kernel dispatch automatically. LLMs assisted in improving the writing of this manuscript.

\section*{Acknowledgment}
We thank Google colleagues Zixiang Zhou, Aahil Mehta, Yaohui Guo, Xiang Zhai, Wen Tang, and Eugene Ie for their technical advice and helpful discussions. We also thank the anonymous reviewers for their constructive feedback.

\end{document}